\documentclass{article}

\usepackage{PRIMEarxiv}
\usepackage{amsthm}
\usepackage{algorithm}
\usepackage{url}
\usepackage{multirow}
\usepackage{amssymb}
\usepackage{subfig}
\usepackage{algorithm}
\usepackage[noend]{algpseudocode}
\usepackage{amsfonts}
\usepackage{filecontents}
\usepackage{mathtools}

\usepackage{xcolor}
\definecolor{rv1}{rgb}{1.0, 0.44, 0.37}
\definecolor{rv2}{rgb}{0.4, 1.0, 0.0}
\definecolor{rv3}{rgb}{0.0, 0.75, 1.0}
\definecolor{rvt}{rgb}{0.75, 0.75, 0.75}
\usepackage{soul}

\newtheoremstyle{exampstyle}
{7pt} 
{7pt} 
{\itshape} 
{} 
{\bfseries} 
{.} 
{.5em} 
{} 
\theoremstyle{exampstyle}

\algrenewcommand\algorithmicindent{2em}%

\usepackage[utf8]{inputenc} 
\usepackage[T1]{fontenc}    
\usepackage{hyperref}       
\usepackage{url}            
\usepackage{booktabs}       
\usepackage{amsfonts}       
\usepackage{nicefrac}       
\usepackage{microtype}      
\usepackage{lipsum}
\usepackage{fancyhdr}       
\usepackage{graphicx}       
\graphicspath{{media/}}     
\usepackage{xspace}
\title{DQLAP: Deep Q-Learning Recommender
Algorithm with Update Policy for a Real Steam
Turbine System
}

\author{
  M.H. Modirrousta\\
  Fault detection and Identification LAB (FDI) \\
  K.N.Toosi University of Technology \\
  Tehran, Iran\\
  \texttt{mohammadbc@email.kntu.ac.ir} \\
  \And
  M. Aliyari Shoorehdeli\\
  Faculty of Electrical Engineering \\
  K.N.Toosi University of Technology \\
  Tehran, Iran\\
  \texttt{aliyari@kntu.ac.ir} \\
  \And
  M. Yari\\
  Faculty of Mechatronic Engineering \\
  K.N.Toosi University of Technology \\
  Tehran, Iran\\
  \texttt{yari.mostafa@mapnaec.com} \\
  \And
  A. Ghahremani\\
  Faculty of Mechanical Engineering \\
  K.N.Toosi University of Technology \\
  Tehran, Iran\\
  \texttt{ghahremani.arash@mapnaec.com} \\
}

\begin{document}
\maketitle

\begin{abstract}
In modern industrial systems, diagnosing faults in time and using the best methods becomes more and more crucial. It is possible to fail a system or to waste resources if faults are not detected or are detected late. Machine learning and deep learning have proposed various methods for data-based fault diagnosis, and we are looking for the most reliable and practical ones. This paper aims to develop a framework based on deep learning and reinforcement learning for fault detection. We can increase accuracy, overcome data imbalance, and better predict future defects by updating the reinforcement learning policy when new data is received. By implementing this method, we will see an increase of $3\%$ in all evaluation metrics, an improvement in prediction speed, and $3\%$ - $4\%$ in all evaluation metrics compared to typical backpropagation multi-layer neural network prediction with similar parameters.
\end{abstract}


\section{Introduction}
Higher reliability is necessary as industrial systems become increasingly specialized and more costly. Detection and analysis errors may result in a decline in performance or even malfunctioning the measuring equipment. The intelligent production of industry 4.0 is based on the use of various new intelligent technologies \cite{b1}, control \cite{b2}, and quality prediction \cite{b3}.

It uses easy-to-measure variables to identify faults from normal process data during the data-driven analysis of faults. It has been extensively proven that data-driven fault diagnosis methods provide flexibility, simplicity, and low cost for fault diagnosis. Recent years have developed various fault diagnosis methods based on machine learning \cite{b4}, \cite{b5}. Today's modern systems heavily rely on the analysis of data and artificial intelligence (AI). The use of artificial neural networks (ANN), machine learning, deep learning, and fuzzy logic in analyzing data is widespread for monitoring, fault detection, and other management functions. Several artificial intelligence systems have detected and found faults in industrial systems \cite{b6}.

It has been demonstrated that classical ML algorithms generally produce easy-to-understand models with substantial mappings. However, their performance saturates as dataset sizes increase. Digitalization and speed of generating data, coupled with ML algorithms that have limitations in handling large datasets, led to the development of deep learning (DL) architectures \cite{b7}. The Deep Learning architecture is constructed of simple mappings that are general approximators \cite{b8}. A deep learning model replaces traditional handcrafted features with trainable layers, which leads to better performance and avoids saturation when applied to large datasets \cite{b9}.

Due to the dependence of supervised learning algorithms on labeled datasets, its capabilities are limited in the digital domain, where plenty of unlabeled data is accessible. Meanwhile, unsupervised algorithms that train with unlabeled data are more effective for shallow architectures \cite{b10}.

Most processes in the industry run smoothly, and faulty samples are rare in the field. Because of this, normal samples are far more common than faulty samples in the industrial process. There are relatively few faulty samples, making implementing traditional fault diagnosis methods challenging. The fault diagnosis domain defines such problems as imbalances in class. Such a problem has been the subject of numerous research studies.

Various sampling or generating procedures help balance the class distribution in the data preprocessing phase. In general, undersampling \cite{b11}, and oversampling \cite{b12} often lead to overfitting and underfitting without guidance or indication. Data generation \cite{b13} can also be unreliable from being practical since it is a novel data-level technique. In addition, Several conditions must be met for Cleaning-Resampling \cite{b14}. There are many domains in which cost-sensitive learning can be applied \cite{b15}.
Nevertheless, it requires experts in the domain to provide the cost matrix in the early stages, which is rarely possible. As a result of developing new loss functions, many recent algorithm-level approaches have been proposed, such as FocalLoss \cite{b16}. The class-imbalance problem in the fault diagnosis domain can be effectively addressed with hybrid approaches. Hybrid approaches are the combination of data-level techniques and algorithm-level methods \cite{b17}.  

All imbalanced industrial process datasets cannot be treated with these methods. They are also sensitive to outliers, which makes their performance fluctuate. Consequently, usability is poor due to the requirement of technical expertise in the design of the cost matrix. In other words, these methods cannot be adapted to complex processes without expert knowledge and presumptions, so they are not universal and not adaptable.

A strategic approach is needed to mitigate the above problems. Furthermore, due to human error, the labels applied by experts to the data from an actual steam turbine may not be reliable. Moreover, a less-knowledgeable algorithm will be used since the label of the data is uncertain. Reinforcement learning algorithms may solve this problem.

Based on the reward-based sequential decision-making process, RL is a branch of machine learning that efficiently and automatically learns and adapts to the environment to find the optimal response to any changes. It is essential to understand that the recommendation process in RL-based recommendation systems is treated as a time-based dynamic interaction between the user and the recommendation agent. As soon as the recommendation system recommends an item to a user, a positive reward will be assigned if the user expresses an interest in it (through clicking or viewing, for example) \cite{b18}.

Using supervised and unsupervised algorithms may expose the user to many problems. In addition, many of these algorithms have difficulties solving unbalanced data problems. Additionally, we are dealing with data that has uncertainty in the labels offered by experts. Due to all these reasons, we designed a reinforcement learning-based recommender system. We have also considered updating the reinforcement learning policy to address the imbalance and uncertainty in labels. A daily data collection schedule for the steam turbine was followed, and the data collection days were indicated. In order to reach an optimal policy, we begin with analyzing the labels from the first day's data, then update the algorithm again with the data of the following day's data. The discussion will continue until the last day, after which we will come up with a conclusion.

In this paper, we describe the main contributions of our framework, referred to as DQLAP:
\begin{itemize}

\item {Building a recommender system using reinforcement learning. Imbalances in data will be handled by this method. Additionally, some problems mentioned with supervised and unsupervised methods do not appear. Our approach enables the expert to make informed decisions without relying on feature engineering.}
\item {Considering the property of transferability and the regularly updated policy, it can give accurate performance based on less data and provide a forecast for the upcoming day.}   
\item {Analyzing the system's performance by comparing it to the declared labels by an independent expert; Since the declared labels cannot be relied upon.} 

    \end{itemize}

\section{Prior Art}
Machine learning (ML) is an emerging approach to fault diagnosis that utilizes artificial intelligence (AI). In fault diagnosis, artificial neural networks (ANN) \cite{b19}, \cite{b20}, support vector machines (SVM) \cite{b21}, and extreme learning machines (ELMs) \cite{b22} have become widely used and effective. In terms of fault detection, these traditional approaches have some limitations. The statistical significance of a fault requires more examples to be collected for it to be significant. This is because few new examples can only do so marginally. A fault that occurs at an early stage can also be challenging to understand because there is a lack of accurate data. Obtaining several valid fault data within a short timeframe is difficult because faults are complex, unstable, and unpredictable. In order to diagnose faults, they rely on hand-crafted feature extractors \cite{b23} that obtain some time- and frequency-domain features.

In the field of machine fault diagnosis, deep learning (DL) \cite{b7}, which has a strong ability to learn features, has gained significant consideration recently \cite{b24}, \cite{b25}. With limited sampling data, some knowledge-driven methods may also be applied when minority samples are scarce. Li et al. \cite{b26} propose a fast and accurate few-shot bearing fault diagnosis method (MLFD) through meta-learning. Zhuo et al. \cite{b27} present a generative model with fault attribute space using the auxiliary triplet loss.

The model must be trained with a large amount of data to reach high accuracy in deep learning. A low level of accuracy or insufficient labeled data reduces the performance of supervised learning.

Deep reinforcement learning (DRL) integrates the capabilities of DL \cite{b7} with the decision-making abilities of reinforcement learning (RL) \cite{b28}. It has achieved great success in gaming, control, and interaction systems \cite{b29}, \cite{b30}, \cite{b31}, \cite{b32}, \cite{b33}. DRL is rarely used in classification tasks because its approach attempts to deal with the sequential decision problem. Using the classification Markov decision process (CMDP), Wiering et al. Described a classification problem as a sequential decision-making process. The resulting MLP network was superior to a typical backpropagation MLP network. With DRL, Lin et al. resolved imbalanced classification by converting imbalanced classification into a sequenced decision-making problem \cite{b34}. Fan et al. designed the DiagSelect framework to perform intelligent imbalanced sample selection using RL to obtain better diagnosis performance autonomously \cite{b35}. Wang et al. outlined a new methodology for fault diagnosis based on time-frequency representations (TFR) and dynamic response mappings (DRLs) \cite{b36}.

Our insights from these successful experiments suggest that we should investigate a comprehensive approach from the viewpoint of DRL as a way to solve the previously mentioned shortcomings of fault diagnosis methods. 

\section{Background}

\subsection{Reinforcement Learning} \label{ssec:RL}
As a reward-based system, reinforcement learning (RL) strives to maximize the rewards of the interactions between an agent and its environment \cite{b28}. In response to feedback from the environment, the agent learns about its behavior, which is then attempted to improve due to its actions. Reward learning problems can be solved by devising policies (e.g., mapping between states and actions) that maximize the accumulation of rewards. In a reinforcement learning problem, five important entities are involved: the state, the action, the reward, the policy, and the value. Generally, reinforcement learning problems are modeled with Markov decision processes.

\subsection{Q-Learning} \label{sssec:QL}
During reinforcement learning, agents learn policies as they transition between states using the Q-learning algorithm \cite{b37}. In order to determine the optimal set of policies, we must assess all possible actions related to the different states of the agent. The value of this algorithm is maintained continuously by updating the Q-value according to the next state of the algorithm and the greedy action. A Q-function is essentially a function that accepts various arguments, such as state vectors, action vectors, rewards vectors, and learning rates. The discount factor is then calculated for the Q-value. However, because Q-learning-based systems require high dimensionality, they do not perform well in large state spaces.

\subsection{Deep Q Network (DQN)}
\label{sssec:DQN}
Several techniques have been developed to deal with significant space state problems, including Deep Q-Network (DQN), a networked Q-learning algorithm combining reinforcement learning with a class of artificial neural networks known as Deep Q Networks. Depending on the outcome of an action, the environment grants a positive reward or penalizes it with a negative reward. In addition to updating the weights of the DNN, the reward is also used to improve the performance of this machine learning algorithm. 

In practice, when playing Atari games, the DQN algorithm has shown to be able to achieve impressive results. The purpose of this algorithm is to integrate DNNs with Q-Learning algorithms to determine self-aware decision policies ${\pi}$ used for mapping relations between states ${S}$ and actions  ${A}$ such that ${A=\pi(S)}$ \cite{b38}.

\subsection{Formulation}
\begin{figure*}
\centering

  \includegraphics[clip, trim=1cm 20cm 0cm 2cm,width=0.7\textwidth,height=4cm, scale=0.2]{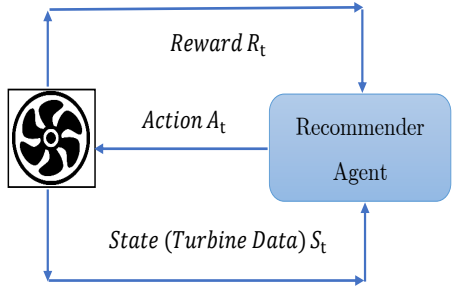}
  \caption{ An overview of the Recommender Agent-User interaction in MDP.}\label{fig:SSL}
\end{figure*}

As shown in Fig \ref{fig:SSL}, the Markov decision process has ${S}$ representing all possible states and ${A}$ representing all possible actions. It is possible to retrieve the possible rewards from any state and action pair ${(s, a)}$ by using the distribution ${R}$. In the Markov decision process, ${P}$ represents the transition probability distribution for the next state, and  ${\gamma}$ represents the discount factor. Future rewards are affected (discounted) by the discount factor. Since an agent faces uncertainty during the process, the more rewards it counts as it tries to predict future outcomes. We want to determine which action should be taken in every state according to a policy that determines the action to take. Furthermore, the complexity of the problem influences the number of states (which affects uncertainty), so our policy function will be complex.  

RL and DRL algorithms were proposed for learning policies with the maximum cumulative reward. In order to determine the optimal policy, we need to: \begin{equation}
\pi^*=\arg \max _\pi \mathbb{E}\left[\sum_{t>0} r^{t r_t \mid \pi}\right]
\end{equation}
Using ${\pi^*}$ as the optimal policy, ${r}$ as rewards, and ${t}$ as the time step.

Value functions represent the expected cumulative reward of being in a particular environment state. It typically refers to the quality of a state and is defined as:  
\begin{equation}
V^\pi(s)=\mathbb{E}\left[\sum_{t \geq 0} r^{t r_t \mid s_0=s, \pi}\right]
\end{equation}
A value function is defined by ${V^\pi(s)}$ based on policy ${\pi}$ for ${s}$, and an expected cumulative reward value is defined by ${\mathbb{E}}$.  

We take action and state into account when calculating the Q-value function. In this case, the expected cumulative reward is calculated by being in a specific state, taking a specific action, and following the policy ${\pi}$.  
\begin{equation}
Q^\pi(s, a)=\mathbb{E}\left[\sum_{t \geq 0} r^{t r_t \mid s_0=s, a_0=a, \pi}\right]
\end{equation}
In this case, ${Q^\pi(s, a)}$ represents the Q-value, ${a}$ represents the chosen action from among all the possible actions (${A}$), and ${s}$ represents a specific state from among all the possible states (${S}$). 

We can retrieve the greatest expected reward from a particular state and action by using ${Q^*(s, a)}$ as the optimal Q-value.  
\begin{equation}
Q^*(s, a)=\max _\pi \mathbb{E}\left[\sum_{t \geq 0} r^{t r_t \mid s_0=s, a_0=a, \pi}\right]
\end{equation}

In the case where the optimal state-action values are identified (${Q^*\left(s^{\prime}, a^{\prime}\right)}$), ${Q^*}$ satisfies the Bellman equation at the next time step:  \begin{equation}
Q^*(s, a)=\mathbb{E}_{s^{\prime} \sim \varepsilon}\left[r+\gamma \max _{a^{\prime}} Q^*\left(s^{\prime}, a^{\prime}\right) \mid s, a\right]
\end{equation}
${\gamma}$ is the discount factor, and ${\varepsilon}$ represents the MDP environment. 

The most effective strategy is to maximize the expected value. The agent uses optimal policy ${\pi^*}$ to decide how to proceed in any state by the specification of ${Q^*}$. Using the Bellman equation as an iteration update, we can find the optimal policy through a value iteration algorithm: \begin{equation}
Q_{i+1}(s, a)=\mathbb{E}\left[r+\gamma \max _{a^{\prime}} Q_i\left(s^{\prime}, a^{\prime}\right) \mid s, a\right]
\end{equation} 

Based on the two equations above, we can rewrite ${Q^*(s, a)}$ as follows:
\begin{equation}
Q^*(s, a)=\mathbb{E}\left[r+\gamma \max _{a^{\prime}} Q_i\left(s^{\prime}, a^{\prime}\right) \mid s, a\right]
\end{equation} 

When neural networks are used in DQN, we get the action-value function ${Q^{\pi}(s, a)}$  from the policy ${\pi}$.
\begin{equation}
    Q(s, a ; \boldsymbol{\theta}) \approx Q^{*}(s, a)
\end{equation}
It is assumed that ${\theta}$ denotes the model parameters. The optimal Q-values are estimated using these weights.  

Agents interact with the environment through greedy algorithms. This algorithm provides a means for balancing "exploration" (random action $a\in A$) with "exploitation" (action $a$ with ${\rm{argma}}{{\rm{x}}_a}\,Q(s,a;\theta )$) and generating a series of experiences defined as $(s_t,a_t,r_t,s_{t+1})$. ER buffer $B$ stores these experience samples. A random sample of experience samples is then taken from ER buffer, which creates a new model based on the randomly generated experience samples.
  
This initial learning stage is accelerated by a dynamic-greedy algorithm generating action distributions to obtain various experience samples in ER buffer $B$, which are stored in it. Changes in parameters can be obtained by following these steps:  
\begin{equation}
\epsilon=\left\{\begin{array}{l}
\epsilon \times \epsilon_{\text {decay }}, \text { if } \epsilon \geq \epsilon_{\min } \\
\epsilon_{\min }, \text { otherwise }
\end{array}\right.
\end{equation}

Q-values can be estimated with an error when using a neural network. Due to this fact, the Bellman equation's error should be minimized. It is possible to minimize the error of the Bellman equation by using training loss functions. As a result, it is a measure of how far ${Q^(s, a)}$ is from the target value ${y_{i}}$ as follows:  
\begin{equation}
L_{i}\left(\boldsymbol{\theta}_{i}\right)=\mathbb{E}\left[\left(y_{i}-Q\left(s, a ; \boldsymbol{\theta}_{i}\right)\right)^{2}\right]
\end{equation}
The model's parameters are updated using ${L_{i}}$ as an objective function representing the mean-square error (MSE). Based on the principle that ${y}$ is the target ${Q}$ value function, the following formula can be used to obtain it: 
\begin{equation}
y_{i}=r+\gamma \max _{a^{\prime}} Q\left(s^{\prime}, a^{\prime} ; \boldsymbol{\theta}_{i-1}\right)
\end{equation}

The optimal policy is determined by updating the objective function using gradient descent. Accordingly, the gradient updating is calculated as follows: 
\begin{equation}
\nabla_{\boldsymbol{\theta}_{i}} L\left(\boldsymbol{\theta}_{i}\right)=\mathbb{E}_{\sim U}\left[\left(y-Q\left(s, a ; \boldsymbol{\theta}_{i}\right)\right) \nabla_{\boldsymbol{\theta}_{i}} Q\left(s, a ; \boldsymbol{\theta}_{i}\right)\right]
\end{equation}
${\sim U}$ indicates the random uniform sampling.
\subsection{Double Deep Q Network (DDQN)} \label{sssec:DDQN}
There are many similarities between DDQN and DQN. The difference between DDQN and DQN lies in that DDQN employs two neural networks, both of which implement current Q functions and the other which implements target Q functions. It is essential to realize that the target Q function is a copy of the current Q function with a delayed timing loop created after a certain number of training runs are completed. 

It is possible to perform gradient descent over objective functions without recursively relying on training networks, thus preventing the moving target effect from occurring in gradient descent \cite{b39}.

We use this approach in the following form:
\begin{equation}
Q^*(s, a)=(1-\alpha) \cdot Q(s, a)+\alpha\left(r+\gamma \cdot \max _{a^{\prime}} Q\left(s^{\prime}, a^{\prime}\right)\right)
\end{equation}
Where $\alpha$ is the learning rate, $\gamma$ is the discount factor, and $\max _{a^{\prime}} Q\left(s^{\prime}, a^{\prime}\right)$ is the estimate of optimal future value.
\subsection{Classification Markov Decision Process (CMDP)}
Because DRL solves a sequential decision-making problem, it is infrequently used for classification tasks. It has been developed that a standard classification problem is defined as a sequential decision-making problem by the classification Markov decision process (CMDP).

Consider the CMDP framework for the tuple $\{S, A, P, R\}$. Classification problems can be described as sequential decision-making problems in which${S}$, ${A}$, ${P}$, and ${R}$ correspond to state space, action space, transition probability, and reward function, respectively.

\textbf{State space ${S}$.} Training samples determine it. In order to determine the initial state of the agent, a sample will be taken at the beginning of each episode. The training data sequence is generated based on the order of samples collected when the environment starts a new episode.

\textbf{Action space ${A}$.} A list of identified actions is provided, and each action is classified as either an attack or a normal label. Consequently,  ${A}$ consists of  $\{1,2, \ldots, K\}$, whereas ${K}$ represents the entire data set.

\textbf{Transition probability ${P}$.} Transition models are described as follows: $P: S \times A \longrightarrow S$. CMDP has a deterministic transition probability. The agent will receive the next state according to the order in which the samples are received in the mini-batch after obtaining the current state and taking the appropriate action. A mini-batch of samples is then processed sequentially until all samples' problems are resolved.

\textbf{Reward function ${R}$.} Rewarding the agent for its actions promotes the agent's learning of the policy and enables it to make appropriate decisions. If the state label is correctly detected, this work results in a positive reward from the environment; otherwise, a negative reward. Therefore, the reward function is designed as follows: 
\begin{equation}
R\left(s_{t}, a_{t}, l_{t}\right)= \begin{cases}1, & \text a_{t}=l_{t} \\ -1, & \text  a_{t} \neq l_{t}\end{cases}
\end{equation}
$l_{t}$ indicates whether the label is a fault or a normal label.

\section{Proposed Method}
We discussed reinforcement learning concepts in the previous section. Our problem is first defined in CMDP format. We use the DDQN algorithm to update our policy, which performs better than DQN.

In order to overcome the problems in the first section, we update our policy daily. There is the capability of collecting data daily for the steam turbine system, and in the data file provided, each line indicates the day the data were collected. From day one, we begin training the algorithm. As a first step, we obtain the model and policy for the first day's data. By using the neural network model's transferability property, we then add the data of the second day to the first day and update the model and policy with the new data. This process continues until the last day (124 days). Algorithm \ref{alg:cap} contains a detailed illustration of the learning phase and the simulated environment model.

\begin{figure*}
  \includegraphics[clip, trim=0cm 12cm 0cm 0.5cm,width=\textwidth, scale=0.02]{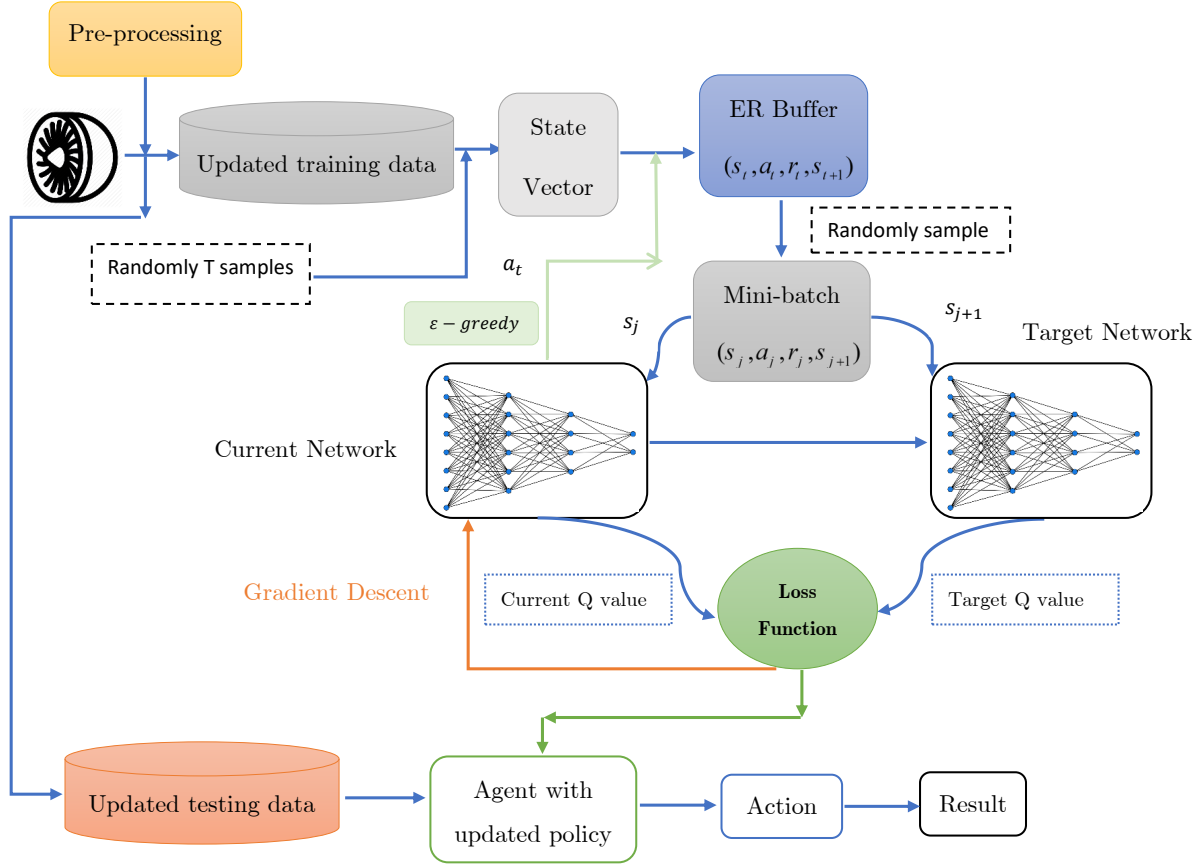}
  \caption{Proposed DRL-based fault detection framework (DQLAP) for steam turbine. }\label{fig:frm}
\end{figure*}

Interacting with the environment will progressively teach the agent an optimal fault detection policy, allowing it to detect faults more precisely. The DQLAP framework is shown in Fig \ref{fig:frm}.  

\begin{algorithm}
\caption{Proposed Double Deep Q-Learning}\label{alg:cap}
\begin{algorithmic}[1]
\Require
        \State States $\mathcal{S} = \{1, \dots, x_n\}$
        \State Actions $\mathcal{A} = \{1, \dots, a_n\},\qquad A: \mathcal{S} \Rightarrow \mathcal{A}$
        \State Reward function $R: \mathcal{S} \times \mathcal{A} \rightarrow \mathbb{R}$
        \State Probabilistic transition function $P: \mathcal{S} \times \mathcal{A} \rightarrow \mathcal{S}$
        \State Learning rate $\alpha \in [0, 1]$, typically $\alpha = 0.001$
        \State Discounting factor $\gamma \in [0, 1]$
        \State Initialize ER buffer $B$ with experience replay memory $M$
        \State Initialize current netwrok $Q_{\theta}$
        \State Initialize target netwrok $Q_{\theta^{\prime}}$
        \Procedure{Double Deep QLearning} {$\mathcal{S}$, $A$, $R$, $T$, $\alpha$, $\gamma$}
\State Store updated data in $D$        
\State \textbf{input:} $D=\left\{\left(x_1, l_1\right),\left(x_2, l_2\right) \ldots \ldots \ldots\left(x_T, l_T\right)\right\}$
\For {each iteration}
\State Shuffle trainind data 
\State Initialize state $s_1 = x_1$
\EndFor
\For {sampled minibatch $\{x_k\}_{k=1}^{N}$}
\EndFor
\For{all $k \in \{1,...,N$ \}}
\State Calculate $\pi$ according to Q and 
\Statex \hspace{20mm} exploration strategy (e.g. $\pi(s) \gets Q(s, a)$)
\State $a \gets \pi(s)$
\State $r \gets R(s, a)$ \Comment{Receive the reward}
\State $s' \gets T(s, a)$ \Comment{Receive the new state}   \State Store $(s_t,a_t,r_t,s_{t+1})$ in $B$
\State Random sample $(s_j,a_j,r_j,s_{j+1})$ from $M$
\State Compute target $Q$ value:
\Statex \hspace{29mm} $
{Q^*}\left( {{s_j},{a_j}} \right) = (1 - \alpha )\,{Q_\theta }\left( {{s_j},{a_j}} \right)\, + \alpha \left( {{r_j} 
 + \gamma {Q_\theta }\left( {{s_{j + 1}},{{{\mathop{\rm argmax}\nolimits} }_{{a^\prime }}}{Q_{{\theta ^\prime }}}\left( {{s_{j + 1}},{a^\prime }} \right)} \right)} \right)$ 
\State $L\left(\boldsymbol{\theta}\right)=(Q^*\left(s_j, a_j\right)-Q_\theta\left(s_j, a_j\right))^2$      
\State Perform gradient descent on $L\left(\boldsymbol{\theta}\right)$ 
\State Update current network parameters
\State Update $\epsilon$
\State Update target network parameters
\EndFor
 \EndProcedure
\end{algorithmic}
\end{algorithm}

\section{Simulations}
This section examines the performance of the proposed fault detection approach.

\begin{table}[!htb]
	\caption{The structure of neural network in DDQN}\label{tab:reg}
	\renewcommand{\arraystretch}{1}
	\label{tab:reg}
	\centering
	\resizebox{0.35\textwidth}{!}{%
		\setlength\tabcolsep{3pt}
		\begin{tabular}{cccccccccc}
		     \hline
			Block&Layers&Neurons\\\hline
			Hidden Layers &Dense & 32\\
			&Dense & 32 \\
			&Dense& 24\\
			&Dense& 24 \\
			&Dense&16 \\
			&Dense&8  \\
			&Dense&4 \\
			\hline
			Classification Head&Linear& 2  \\
			 \hline
			Parameters && 4054\\
			\hline
		\end{tabular}
	}
\end{table}

\subsection{Setup Details}

\textbf{Dataset.}\label{AA}
Data from an actual working steam turbine was used in this study. A sampling rate of one minute per day was used to collect the turbine's data over 124 days. It was determined that 207,361 data were collected in total. However, 190,635 data could be used after preprocessing and removing outlier data from the non-dominant operating mode. The normal category comprises 121,279 data, and the fault category comprises 69,356 data. It contains 31 features, including condenser pressure, pressure on inlet valves, steam flow value, active and reactive power, and others. In this case, the problem was caused by the leakage of one of the pressure valves, failing the entire system. Data are collected in the dominant operating mode of the system, known as the high-pressure mode, and they are sufficiently comprehensive to provide a good understanding of the system's functioning.

\textbf{Parameters.}\label{AA}
With the CMDP framework, we set the mini-batch size to 128, initial $\alpha$ to 1 with a decay rate of 0.9999, and initial $\epsilon$ to 0.9 with a decay rate of 0.99. Additionally, we set the $\gamma$ to 0.001 and the number of iterations to 1000, with the minimum $\alpha$ and $\epsilon$ setting to 0.0001. As part of the training process, we also use an Adam optimizer with learning rates of 0.001. Table 1 shows the parameters of the neural networks, and both the current and target networks have the same structure. In hidden neurons, we use the LeakyReLU activation function. There is no reason to worry about initializing a neural network with an LReLU. It is possible for a ReLU network never to learn if neurons are not activated at the beginning.

At each step of the algorithm execution, $70\%$ of the data is used for training, while the rest is used to test the algorithm. Every time we train the algorithm on daily data, we return a higher-performing model which returns the highest number of recall criteria in all iterations (on test data) to increase the speed of analysis. In order to demonstrate the importance of the class with less data, we consider average recall as a macro.

\textbf{Preprocessing.}\label{AA}
In the available dataset, some rows have a "drop-it" title. Data such as this should be deleted as it is considered outlier data. It is then necessary to normalize the data between 0 and 1 by using a Min-Max scaler as follows:

\begin{align}
{{\rm{x}}_{scaled}} = \frac{{{\rm{x}} - {\rm{ }}{{\rm{x}}_{min}}}}{{{{\rm{x}}_{max}} - {\rm{ }}{{\rm{x}}_{min}}}}.
\label{formula3}
\end{align}
Where $x$ and $x_{scaled}$ assigns features before and after normalization, and $x_{min}$ and $x_{max}$ assigns minimum and maximum value of features based on training data before normalization.

\subsection{Results and Discussion}
We analyze the dataset for all days at once with the framework specifications DQLAP and the parameters mentioned in the previous section. According to the results, we achieved $91\%$ in all performance metrics. According to this performance level, the deep reinforcement learning algorithm's analysis is close to that of an expert. 

By using the daily update to improve the policy instead of all the data at once, we reach $94\%$ in all performance metrics. As a result, the daily update policy is more closely related to the expert, and the algorithm is more similar to the expert's daily feedback. 

Our results show that the neural network does not perform as well as the previous two models outside the designed framework. 

Furthermore, the prediction speed of the network in the framework (per iteration) is faster than that of the same network outside it. Our analysis can be sped up by using a buffer and creating action parameters that interact with the environment.
\begin{table} [!htb]
	\caption{Comparison of designed systems}\label{tab:transfer}
	\renewcommand{\arraystretch}{1.5}
	\label{tab:reg}
	\centering
	\resizebox{0.48\textwidth}{!}{%
		\setlength\tabcolsep{3pt}
		\begin{tabular}{cccccccccc}
			\hline
			method & Accuracy& Precision & Recall& F1 Score \\ \hline
			MLP Network  &0.90&0.91&0.90&0.90\\
			DDQN without Update Policy  &0.91&0.91&0.91&0.91\\
			DDQN with Update Policy &\textbf{0.94}&\textbf{0.94}&\textbf{0.94}&\textbf{0.94}\\
			\hline		
			
		\end{tabular}
	}
\end{table}
\subsection{Visualization}
\begin{figure}[H]
\centering

  \includegraphics[scale=0.6]{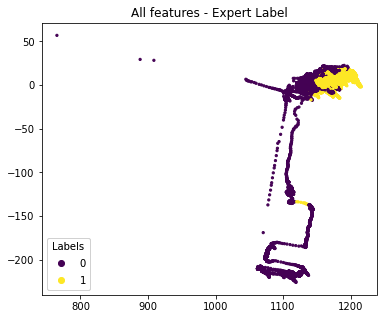}
  \caption{Scatter plot for two-dimensional representation in the spatial distribution of data with the label suggested by the expert.}\label{fig:v1}
\end{figure}

\begin{figure}[H]
\centering

  \includegraphics[scale=0.6]{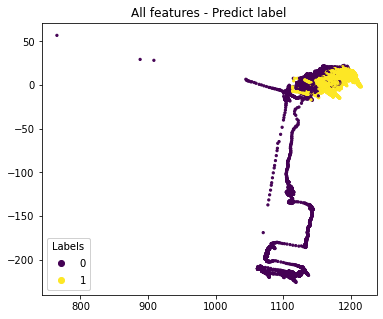}
  \caption{Scatter plot for two-dimensional representation in the spatial distribution of data with the label suggested by our framework.}\label{fig:v2}
\end{figure}

We map the entire data to 2D space using PCA to understand the recommender system's performance and compare it with the announced labels. This will allow us to analyze the data visually. Fig \ref{fig:v1} and \ref{fig:v2} depict this representation. Fault data is labeled 0, and normal data is labeled 1. The expert and DQLAP labels are the same in most distribution areas, as shown in Fig \ref{fig:v1} and Fig \ref{fig:v2}. There is evidence that DQLAP has declared more normal labels in the above cluster and that the distribution of those areas has become more coherent. By improving our framework, we can improve this cluster's coherence and classification accuracy.

 As seen in Fig \ref{fig:v1}, there is a yellow part of the data distribution in the middle of the page, indicating that the expert declared that area normal. The data distribution seems closer to non-normal or fault, and DQLAP has also categorized it as a fault.

Consequently, the differences between the expert's opinion and the prediction framework appear primarily due to the closeness of the distributions of the two classes. DQLAP's division form seemed better than the expert opinion in some distribution parts.

\section{Conclusion}
This paper presents a recommender framework combining reinforcement learning with a turbine system's neural networks. To solve our problem, we first defined it as a Markov Decision Process, then analyzed and compared it with the feedback of an expert by obtaining suitable parameters for the network and the reinforcement learning agent. The results show that the proposed algorithm has the closest performance to an expert's opinion if we analyze the data daily and update the policy accordingly. Furthermore, our system will perform the best analysis with less label knowledge by learning the distribution space of data and, based on that, announce its forecast for the next few days to help expert man and reduce time waste, and cost, along with detecting faults that are unlikely to occur. Moreover, this method does not require extracting features. DQLAP is better at predicting each iteration than a similar network outside the framework, and the analysis can be performed at more speed every day.

\bibliographystyle{unsrt}  
\bibliography{references}

\end{document}